\documentclass[letterpaper, 10 pt, conference]{ieeeconf}
\IEEEoverridecommandlockouts                                       
\usepackage{siunitx}
\usepackage{config/noel}

\title{\LARGE \bf
Nonlinear Model Predictive Control of a 3D Hopping Robot: \\ Leveraging Lie Group Integrators for Dynamically Stable Behaviors
}
\author{Noel Csomay-Shanklin, Victor D. Dorobantu, and Aaron D. Ames%
\thanks{This research was supported by NSF Graduate Research Fellowship No. DGE‐1745301 and Raytheon, Beyond Limits, JPL RTD 1643049.}\thanks{Authors are with the Department of Computing and Mathematical Sciences, California Institute of Technology,
Pasadena, CA 91125. }%
}

\begin{document}
\maketitle
\thispagestyle{empty}
\pagestyle{empty}



\begin{abstract}
Achieving stable hopping has been a hallmark challenge in the field of dynamic legged locomotion.  Controlled hopping is notably difficult due to extended periods of underactuation combined with very short ground phases wherein ground interactions must be modulated to regulate global state. In this work, we explore the use of hybrid nonlinear model predictive control paired with a low-level feedback controller in a multi-rate hierarchy to achieve dynamically stable motions on a 3D hopping robot. In order to demonstrate richer behaviors on the manifold of rotations, both the planning and feedback layers must be designed in a geometrically consistent fashion; therefore, we develop the necessary tools to employ Lie group integrators and appropriate feedback controllers. We experimentally demonstrate stable 3D hopping, as well as trajectory tracking and flipping in simulation. 

\end{abstract}


\section{Introduction}

Hopping has been a benchmark challenge in the field of robotic locomotion dating back to the seminal work of Marc Raibert in the 1980's \cite{raibert_experiments_1984}.
The lessons learned during hopping have inspired generations of researchers, and have enabled complex behaviors such as walking and running on bipedal \cite{raibert_dynamically_nodate, dadashzadeh_template_2014, reher_dynamic_2021} and quadrupedal robots \cite{poulakakis_modeling_2005, park_high-speed_2017}. Since the original work of Raibert, the landscape of control has changed dramatically, and has recently been fueled by advances in computation power allowing for previously prohibitively costly methodologies to be employed in real-time on robotic platforms. The goal of this work is to investigate the extent to which this new modern computation can be leveraged in the context of the canonical example of hopping. 

The control of hopping robots is particularly challenging due to intermittent continuous and discrete dynamics, periods of extreme underactuation, and exceptionally short ground phases during which the robot can apply forces to regulate its global position. These pose unique difficulties for conventional control algorithms, and necessitate the ability to decide control actions based on predictions of where the robot will be in the future. Raibert addressed these challenges through the use of reduced order models, e.g., the spring loaded inverted pendulum (SLIP) model \cite{geyer_compliant_2006}, which has since proven effective in walking and running generation on bipedal robots \cite{hubicki_atrias_2016}.  Besides the work of Raibert, many methods have been developed to stabilize hopping robots \cite{sayyad_single-legged_2007}, including reinforcement learning based approaches \cite{tedrake_improved_nodate,maier_neural_2001}, nonholonomic motion planning \cite{murray_steering_1990}, a mix of offline and online hierarchical motion planning strategies \cite{zeglin_control_1998,albro_optimal_2001}, and model predictive control of simplified models\cite{zamani_nonlinear_2020, rutschmann_nonlinear_2012}. Compared to the abundance of work that exists for planar hopping robots, the literature for the control of 3D hopping robots is comparatively sparse.

\begin{figure}[t!]
    \centering
    \href{https://youtu.be/0XMxluC_Gs8}{
    \includegraphics[width=0.97\columnwidth]{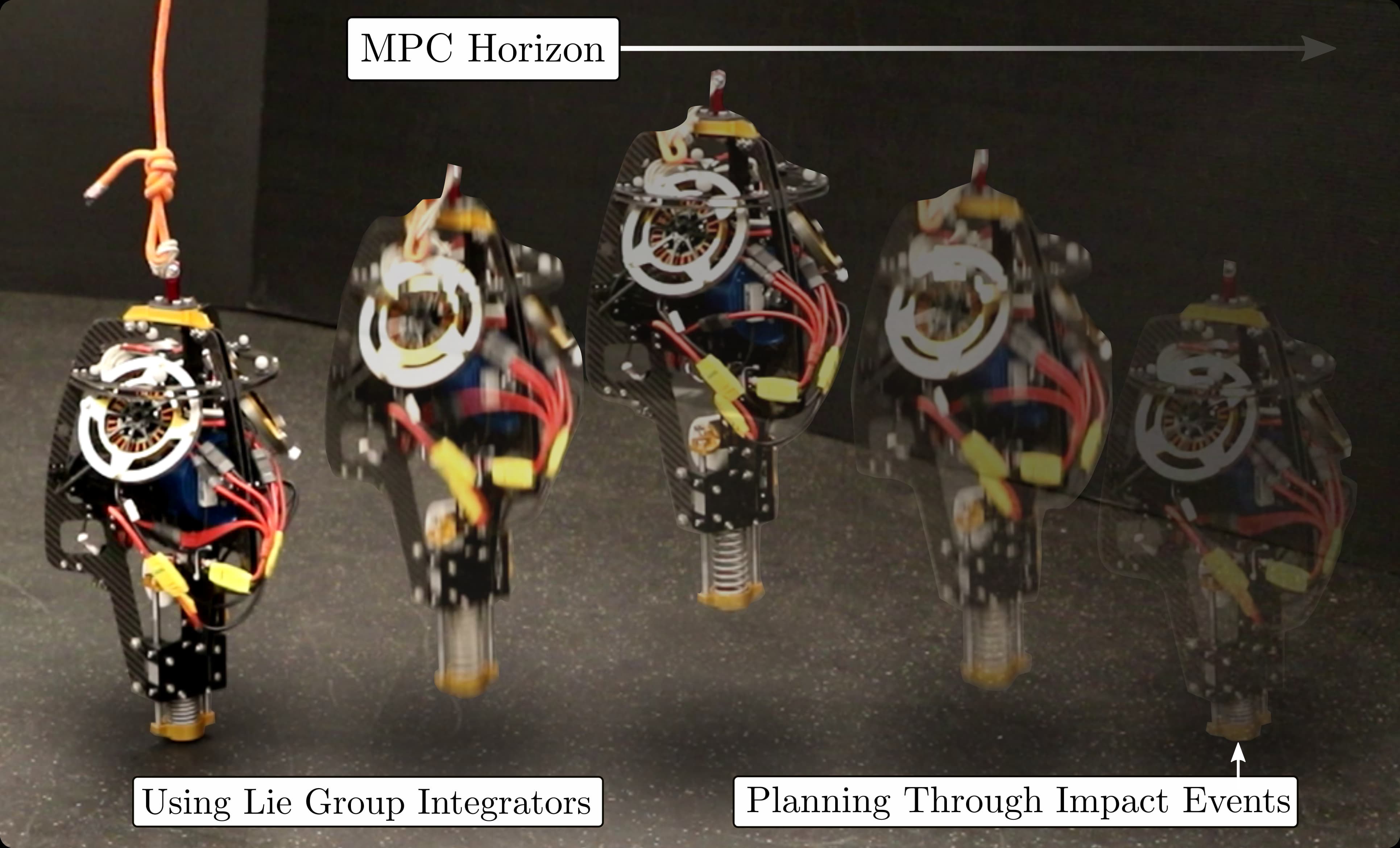}
    }
    \caption{Nonlinear MPC running in real-time on the 3D robot ARCHER \cite{ambrose_creating_2022}, demonstrating hopping on hardware.\vspace{0mm}}
    \label{fig:intro}
\end{figure}

Recently, model predictive control (MPC) \cite{borrelli_predictive_2017} has been used effectively for the control of dynamic robotic systems, including hybrid systems (systems with continuous dynamics and discrete impacts \cite{grizzle_models_2014}) like legged robots \cite{galliker_bipedal_2022, farshidian_real-time_2017, di_carlo_dynamic_2018}. MPC has been brought into the realm of real-time dynamic robotic control due to modern computing power and increased algorithmic efficiency \cite{carpentier_pinocchio_2019}; however, the implementation of MPC on hardware platforms, as well as theoretical justifications of its performance for nonlinear systems remains an active area of research. 
As nonlinear MPC is predicated on taking local approximations of system dynamics, its success heavily relies on correctly constructing these approximations and remaining within the regions in which they are valid.

When the system states are manifold-valued, these local approximations must be carefully constructed. We will specifically be concerned with Lie groups, groups with smooth manifold structure whose operator and inverse are also smooth (see \cite{celledoni2014introduction, kobilarov2009lie}), as they are often used in the field of robotics to model the space of orientations. Lie groups have additionally been studied from the perspectives of discrete mechanics \cite{junge2005discrete} and numerical analysis \cite{iserles2000lie, hairer2006geometric}. The estimation of robotic systems on Lie groups was outlined in \cite{sola_micro_2021}, and the control of legged robotic systems on Lie groups has recently been investigated in \cite{teng_error-state_2022}. Finally, the application of optimal control techniques over differentiable manifolds to the control of hybrid systems was explored in \cite{mastalli20crocoddyl} with experimental results achieved in \cite{mastalli_feasibility-driven_2022, mastalli_inverse-dynamics_2023}. 


The goal of this work is to experimentally realize dynamic hopping behaviors on a 3D hopping robot through the use of nonlinear MPC. To this end, we: (1) develop a framework for hybrid nonlinear MPC in a geometrically consistent fashion via Lie group integrators, and (2) experimentally demonstrate on a 3D robot that MPC is an effective tool for accomplishing highly dynamic behaviors such as hopping.  

Theoretically, we consider both the manifold structure of the configuration space and the hybrid nature of the dynamics (Sec. \ref{sec:preliminaries}), wherein an MPC problem is synthesized through the use of sequential linearizations that leverage Lie group and Lie algebra structures (Sec. \ref{sec:MPC}). The MPC problem is translated to hardware via a multi-rate control paradigm. 
To experimentally verify this framework (Sec. \ref{sec:results}), we leverage a newly built hardware platform: ARCHER \cite{ambrose_creating_2022} (which builds upon earlier generations of hopping robots \cite{ambrose_towards_2021, ambrose_improved_2020, ambrose_design_2019}).  This robot has three reaction wheels for attitude control, and one motor connected via a rope to control foot spring compression.  We first show the capabilities of this robot in a high-fidelity simulation wherein a variety of dynamic motions are demonstrated: from path following to flipping.  Finally, we experimentally realize sustained 3D hopping on this new hardware platform, marking the first demonstration of 3D hopping using online motion planning strategies.

\section{Preliminaries}
\label{sec:preliminaries}

\subsection{Hybrid System Dynamics}
The configuration of the hopping robot is given by $\b q = (\b p, \quat, \b \theta,  \ell)\in\mathcal{Q}$, where $\b p\in\R^3$ is the Cartesian position, $\quat\in S^3$ is the unit quaternion representing the orientation, $\b \theta\in\R^3$ represents the flywheel angles, and $\ell\in\R$ is the foot deflection. 
Next, let $\b v = (\dot{\b p}, \b \omega, \dot{\b \theta},\dot{ \ell})\in \mathcal{V}\triangleq \R^3\times \mathfrak{s}^3\times \R^3\times \R$, where $\bm \omega\in\mathfrak{s}^3$ is a purely imaginary quaternion representing the angular rate of the body.
The complete robot state can then be written as $\b x = (\b q,\b v) \in \mathcal{X} \triangleq \mathcal{Q}\times \mathcal{V}$. 

\begin{figure}
    \centering
    \includegraphics[width=\columnwidth]{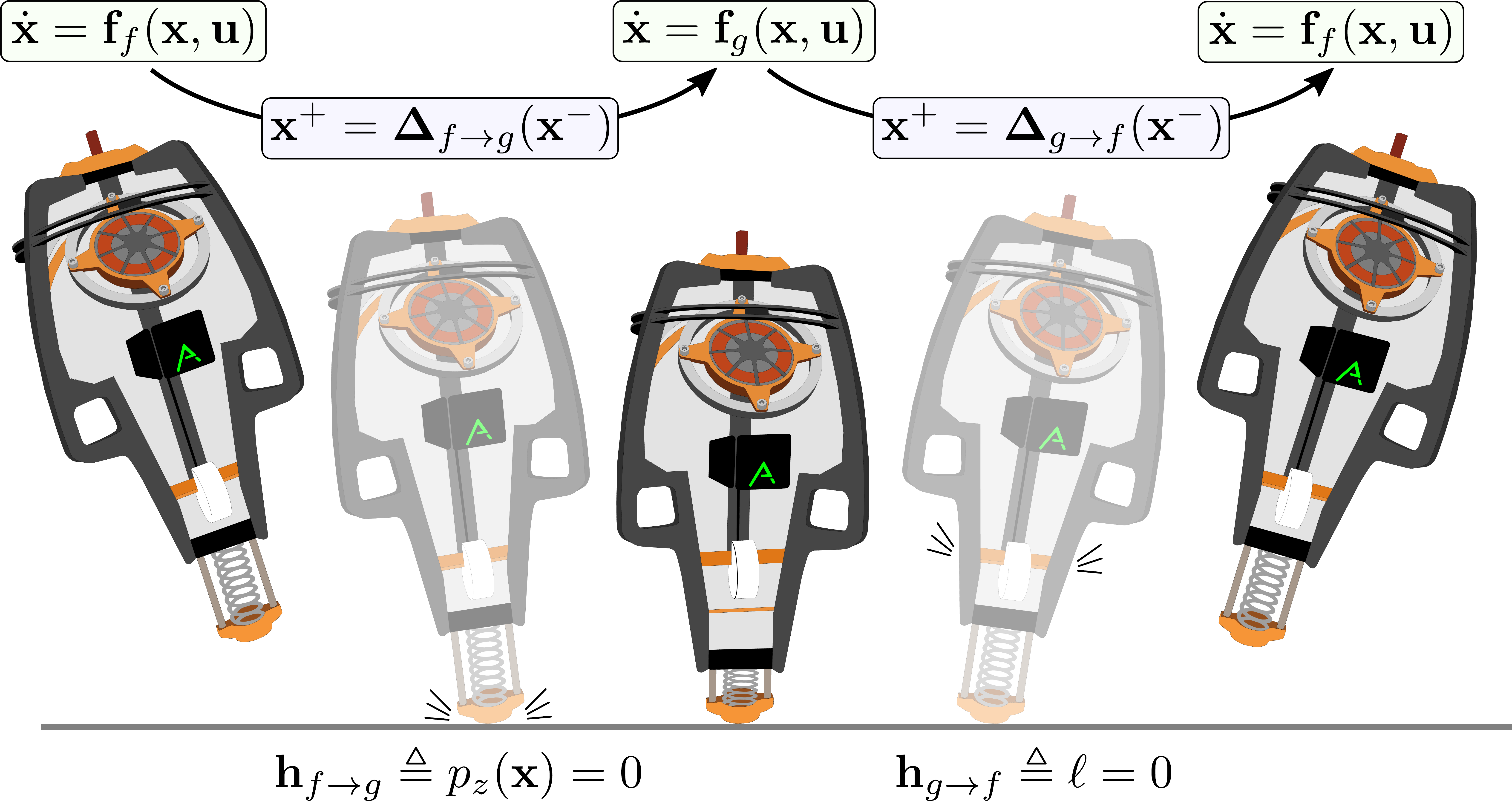}
    \caption{{The robot traversing the various hybrid domains.}}
    \vspace{3mm}
    \label{fig:HybridGraph}
\end{figure}

Hopping consists of alternating sequences of continuous and discrete dynamics; therefore, it is naturally modeled as a hybrid system. Two distinct continuous phases of dynamics exist, the \textit{flight} phase $f$ when the robot is in the air, and the \textit{ground} phase $g$ when the foot is contacting the floor. We can construct a directed graph with vertices $v\in V\triangleq\{f, g\}$ and edges $e \in E\triangleq\{f\to g, g\to f\}$ to characterize how the robot traverses the hybrid modes, as shown in \figref{fig:HybridGraph}. 

For each vertex $v\in V$, let $\mathcal{D}_v\subset\mathcal{X}$ represent the admissible domain in which the system state evolves, and $n_v$ denote the number of holonomic constraints restricting the motion of the robot. Note that $n_f=0$, and $n_g = 3$, which pin the foot to the ground. Omitting the details for manifold value variables (which can be found in \cite{abraham_foundations_2008}), let $\b J_v:\mathcal{Q}\to \R^{n_v \times n}$ denote the Jacobian of the holonomic constraints where $n = |\mathcal{Q}|$. We can define the dynamics as:
\begin{align*}
    \b D(\b q)\ddot{\b q} + \b H(\q, \dot{\q}) &= \b B \b u + \b J^\top_v(\q) \b \lambda_v, \\ 
    \dot{\b J\hspace{1pt}}_v(\q,\dot{\q}) \dot {\q} + \b J_v(\q) \ddot {\q} &= \b 0.
\end{align*}
where $\b D:\mathcal{Q}\to \mathbb{R}^{n\times n}$ is the mass-inertia matrix, $\b H:\mathcal{X}\to \mathbb{R}^n$ contains the Coriolis and gravity terms, $\b B\in\mathbb{R}^{n\times m}$ is the actuation matrix, $\b u \in \mathbb{R}^m$ is the control input, and $\b \lambda_v\in \R^{n_v}$ are the Lagrange variables describing the constraint forces. These equations can be rearranged and forces $\b \lambda_v$ solved for in order to produce the constraint implicit dynamics:
\begin{align}
\label{eqn:nl_dyn}
    \dot{\b x} = \b f_{v}(\b x, \b u),
\end{align}
where $\b f_v:\mathcal{X}\times \mathbb{R}^m\to \mathbb{R}^{2n}$ is of control-affine form. 





Each hybrid transition $e\in E$ occurs when the system state intersects the \textit{guard}, defined as: 
\begin{alignat*}{2}
    \mathcal{S}_{f} &= \{\b x \in \mathcal{D}_{f} : p_z(q) = 0, \dot{p}_z(x) < 0\}~~~&&e = f\to g,\\
    \mathcal{S}_{g} &= \{\b x \in \mathcal{D}_{g} : \ell = 0, \dot{\ell} < 0\}&&e = g\to f,
\end{alignat*}
where $p_z:\mathcal{Q}\to\R$ returns the vertical position of the foot. For an edge $e = v_1\to v_2$, upon striking the guard $\mathcal{S}_{v_1}$ the system undergoes a discrete jump in states as described by:
\begin{align*}
    \b x^+ = \b \Delta_e(\x^-),
\end{align*}
where $\b \Delta_e : \mathcal{S}_{v_1} \to \mathcal{D}_{v_2}$ is the \textit{reset map} describing the momentum transfer though impact, and $\b x^-\in \mathcal{D}_{v_1}$ and $\b x^+\in \mathcal{D}_{v_2}$ are the pre and post-impact states, respectively. Collecting the various objects $\mathcal{D} = \{\mathcal{D}_v\}_{v\in V}$, $\mathcal{S}=\{\mathcal{S}_v\}_{v\in V}$, $\b\Delta = \{\b\Delta_e\}_{e\in E}$ and $F=\{\b f_v\}_{v\in V}$, we can describe the hybrid control system of the hopping robot via the tuple:
\begin{align*}
    \mathcal{HC} = (V, E, \mathcal{D}, \mathcal{S}, \b\Delta, F).
\end{align*}

\begin{figure*}
    \centering
    \includegraphics[width=0.97\textwidth]{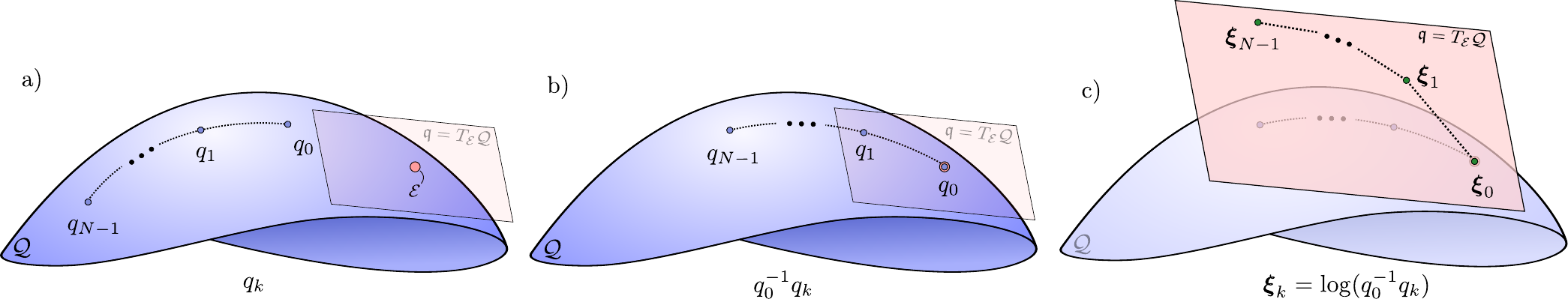}
    \caption{A depiction of Lie groups, Lie algebras, and the $\log$ operation. a) The trajectory ${\quat}_k$, b) pulling the trajectory back to the identity element via ${\quat}_0^{-1}$, and c) taking the $\log$ map near identity to obtain elements in the Lie algebra.\vspace{-4mm}}
    \label{fig:LieGroup}
\end{figure*}



\subsection{Lie Group Integrators}
\label{sec:linearization}
In this section, we focus our attention to the orientation coordinates of the robot and discuss how to perform one form of Lie group integration. 
We represent the orientation of a rigid body via a unit quaternion $\quat\in S^3 = \{\quat\in\mathbb{H}:|\quat| = 1\}$; quaternionic representations of orientation have been extensively explored in attitude control of spacecrafts \cite{kalabic_mpc_2017, yang_spacecraft_2012}. As opposed to Euler angles, quaternions do not suffer from issues of singularities, and provide a straightforward interpolation method -- this will be helpful when constructing continuous-time signals from the discrete points that MPC produces. 
Importantly, quaternions and their associated quaternion multiplication define a Lie group structure on $S^3$. The angular rate of the body is given by an element of the associated Lie algebra $\bm \omega\in\mathfrak{s}^3$, the tangent space of $S^3$ at the identity element $\quat_{\mathcal{E}},$ the elements of which are purely imaginary quaternions.

The time rate of change of a unit quaternion at a point $\quat$ is given by an element of the tangent space at $\quat$, i.e. $\dot \quat \in T_\quat S^3$. Given the angular rate of the body $\b\omega$, we can calculate $\dot \quat$ as:
\begin{align}
    \dot \quat = \quat\b\omega, \label{eqn:q_dot}
\end{align}
using standard quaternion multiplication. The formulation \eqref{eqn:q_dot} is possible since the tangent map of left multiplication by a quaternion is also given by left multiplication, mapping from the Lie algebra to the tangent space at $\quat$.\footnote{Often, the equation \eqref{eqn:q_dot} is reported as $\dot \quat = \frac{1}{2}\quat\b\omega$. This is because the isomorphism between $\b\omega \in \mathfrak{s}^3$ and $\mathbb{R}^3$ is given by $\phi(\b\omega) = \frac{1}{2}\b\omega$ if the generators of $\mathfrak{s}^3$ are taken to be the canonical basis of imaginary 3-vectors, which arises from the fact that the action of quaternions parameterized by a rotation angle $\theta$ on vectors rotates them by an angle of $2\theta$.}
Integrating equation \eqref{eqn:q_dot} results in:
\begin{align}
\label{eqn:lie_int}
    \quat(t) = \quat(0)\exp(\b\omega(t))
\end{align}
where $\exp:\mathfrak{s}^3\to S^3$ is termed the exponential map and maps elements of the Lie algebra $\mathfrak{s}^3$ back to the Lie group $S^3$. This map is injective for imaginary quaternions with magnitude less than $\pi$; over this neighborhood, the inverse map is denoted as $\log:S^3\to\mathfrak{s}^3$. If instead of directly integrating we take a \textit{Lie-Euler step}, \eqref{eqn:lie_int} becomes:
\begin{align}
\label{eqn:lieEuler}
    \quat_{k+1} = \quat_k\exp(\b\omega_k h)
\end{align}
for a (small) time step $h\in\R$. This is a simple example of a \textit{Lie group integrator}.

\section{Geometric MPC and Multi-Rate Control}
\label{sec:MPC}

\subsection{Linearized Dynamics}
In order to avoid the nonlinearity present in \eqref{eqn:lieEuler}, we propose the following change of coordinates:
%
\begin{align}
\label{eqn:xi_def}
    \b\xi_k = \log(\quat_0^{-1}\quat_k),
\end{align}
which first pulls the variables back to the vicinity of $\quat_\varepsilon$, and then to the Lie algebra as shown in \figref{fig:LieGroup}. Substituting in the Lie-Euler step from \eqref{eqn:lieEuler} into the above expression for the first few values of $k$, we have: 
\begin{align*}
    \b\xi_0 &= \b 0 \\ 
    \b\xi_1 &= \log(\quat_0^{-1}\quat_1) = \log(\quat_0^{-1}\quat_0\exp(\b\omega_0 h)) \\
    &= \b\omega_0 h \\
    \b\xi_{2} &= \log(\quat_0^{-1}\quat_2) = \log(\quat_0^{-1}\quat_1\exp(\b\omega_1 h))\\
    &= \log(\quat_0^{-1}\quat_0\exp(\b\omega_0 h)\exp(\b\omega_1 h)) \\
    &= \b\omega_0 h + \b\omega_1 h + \frac{1}{2}[\b\omega_0 h, \b\omega_1 h] + O(h^3)
\end{align*}
where $[\cdot, \cdot]$ represents the Lie bracket, the last line follows from the Campbell-Baker-Hausdorff theorem \cite{stillwell_matrix_2008}, and the higher order terms consist of linear combinations of iterated Lie brackets, which due to linearity are multiplied by terms of $h^3$ or higher\footnote{If quaternion multiplication commuted, then the iterated Lie brackets would be zero -- but alas, no such pleasantries exist.}. This means that, neglecting terms of $O(h^2)$ or higher, we are able to write our dynamics update law as:
\begin{align*}
    \b\xi_{k+1} &= \sum_{i=0}^k \b\omega_k h = \b\xi_k + \b\omega_k h,
\end{align*}
which is linear and will therefore be straightforward to include in the MPC program. 


The next challenge concerns the construction of local approximations of the acceleration dynamics. As most of the coordinates lie in Euclidean space and therefore have straightforward Taylor approximations, we will limit our attention on the manifold valued variables. Specifically, in a continuous domain $v$ consider a function $f:S^3\times \frak{s}^3 \times \R^m \to \frak{s}^3$ satisfying:
\begin{align*}
    \dot{\b\omega} = f(\quat,\b\omega, \b u).
\end{align*}
Given a perturbation $\b\eta\in\frak{s}^3$, a local representation of $f$ in exponential coordinates $\tilde{f}:S^3 \times \mathfrak{s}^3 \times \R^m \times \mathfrak{s}^3\to \R^3$ can be defined as:
\begin{align*}
    \tilde f(\quat, \b\omega, \b u, \b\eta) \triangleq f(\quat\exp(\b\eta), \b\omega, \b u).
\end{align*}
Given $\quat \in S^3$, $\b \omega \in \mathfrak{s}^3$, and $\b u \in \R^m$, as well as additional perturbations $\Delta\b\omega\in\R^3$ and $\Delta \b u \in \R^m$, we can compute a Taylor expansion of $\tilde f$ about the point $(\quat, \b\omega, \b u, \mb 0)$ at a perturbed point $(\quat, \b\omega + \Delta \b\omega, \b u + \Delta\b u, \b \eta)$ via:
\begin{align*}
    &f(\quat\exp(\b\eta), \b\omega + \Delta\b\omega, \b u + \Delta\b u) \\ &\qquad \approx f(\quat, \b\omega, \b u)
     + \derp{\tilde f}{\b{\eta}} \cdot \b\eta + \derp{f}{\b\omega} \cdot \Delta\b\omega + \derp{f}{\b u} \cdot\Delta\b u.
\end{align*}
%
Then, we can write the continuous-time linearized dynamics of $\b\omega$ about the point $(\quat, \b\omega, \b u)$ as:
\begin{align}
\label{eqn:w_dyn}
    \frac{\mathrm{d}}{\mathrm{dt}}{\delta \b\omega} &= \derp{\tilde f}{\b\eta}(\quat, \b\omega, \b u, \b 0)\cdot \b\eta + \derp{f}{\b\omega} \cdot \delta\b\omega + \derp{f}{\b u} \cdot \delta\b u. 
\end{align}

Next, we consider the dynamics of the variables $\b\xi_k$ around a reference trajectory $\bar{\quat}_k \in S^3$, $\bar{\b\omega}_k \in \mathfrak{s}^3$, and $\bar{\b u}_k \in \R^m$. Define $\bar{\b\xi}_k = \log(\bar{\quat}_0^{-1}\bar{\quat}_k)$ with $\bar{\b\xi}_0 = \b 0$ and suppose the reference trajectory satisfies the Lie-Euler step, i.e.:
\begin{equation*}
    \bar{\b\xi}_{k + 1} = \bar{\b\xi}_k + \bar{\b\omega}_k h.
\end{equation*}
For a trajectory $\b\xi_k \in \mathfrak{s}^3$ similarly satisfying the Lie-Euler step, and vectors $\b\omega_k \in \mathfrak{s}^3$, and $\b u_k \in \R^m$, we have:
\begin{equation*}
    (\b\xi_{k + 1} - \bar{\b\xi}_{k + 1}) = (\b\xi_k - \bar{\b\xi}_k) + (\b\omega_k - \bar{\b\omega}_k)h + \underbrace{\bar{\b\xi}_k + \bar{\b\omega}_k h - \bar{\b\xi}_{k + 1}}_{=\b 0},
\end{equation*}
yielding continuous-time linearized dynamics:
\begin{equation}
    \label{eqn:xi_dyn}
    \frac{\mathrm{d}}{\mathrm{dt}}{\delta\b\xi} = \delta\b\omega.
\end{equation}
Combining expressions \eqref{eqn:w_dyn} and \eqref{eqn:xi_dyn}, we obtain:
\begin{equation}
\label{eqn:cont_lin}
    \frac{\mathrm{d}}{\mathrm{d}t} \begin{bmatrix} \delta\b\xi \\ \delta\b\omega \end{bmatrix} = \underbrace{\begin{bmatrix} \b 0 & \b I \\ \derp{\tilde f}{\b\eta}(\quat, \b\omega, \b u, \b 0) & \derp{f}{\b\omega} \end{bmatrix}}_{\b A} \begin{bmatrix} \delta\b\xi \\ \delta\b\omega \end{bmatrix} + \underbrace{\begin{bmatrix} \b 0 \\ \derp{f}{\b u} \end{bmatrix}}_{\b B} \delta\b u.
\end{equation}
%

We can similarly construct local approximations of the impact maps. On an edge $e$, consider a function $\Delta: \mathcal{S} \to \mathfrak{s}^3$ which satisfies:
\begin{equation}
    \b\omega^+ = \Delta(\quat^-, \b\omega^-).
\end{equation}
Here, $\mathcal{S}$ represents the guard as a submanifold of $S^3 \times \mathfrak{s}^3$ (though the guard is actually a submanifold of $\mathcal{Q}$). As defined, this reset map is the restriction of the momentum transfer of the system at impact to the guard. Therefore, we can naturally extend the domain of the reset map by considering the same momentum transfer applied anywhere in the state space, yielding ${\Delta}_{\mathrm{ext}}: S^3 \times \mathfrak{s}^3 \to \mathfrak{s}^3$. This is needed because in our Talyor expansion of the discrete dynamics, we consider perturbations of the full system state (not just perturbations tangent to the guard)\footnote{This extension allows us to abscond from having to only consider perturbations along the guard, which is an interesting area of future work.}. As before, we can locally approximate the function $\Delta_{\mathrm{ext}}$ via:
\begin{align*}
    &\Delta_{\mathrm{ext}}(\quat^-\exp(\b\eta), \b\omega^- + \Delta \b\omega^-) \approx \Delta_{\mathrm{ext}}(\quat^-, \b\omega^-)\\
    &\qquad\qquad\qquad~ + \derp{\tilde{\Delta}_{\mathrm{ext}}}{\b\eta}(\quat^-, \b\omega^-, \b 0) \cdot \b\eta + \derp{\Delta_{\mathrm{ext}}}{\b\omega^-} \cdot \Delta \b\omega^-,
\end{align*}
where again we can represent $\Delta_{\mathrm{ext}}$ locally as:
\begin{align*}
    \tilde{\Delta}_{\mathrm{ext}}(\quat^-, \b\omega^-, \b\eta) \triangleq \Delta_{\mathrm{ext}}(\quat^-\exp(\b\eta), \b\omega^-). 
\end{align*}
Noting that the $\quat^+ = \quat^-$, we can represent the linearization of the discrete map as:
\begin{align}
    \label{eqn:disc_lin}
    \begin{bmatrix} \delta\b\xi^+ \\ \delta\b\omega^+ \end{bmatrix} = \underbrace{\begin{bmatrix} \b I & \b 0 \\ \derp{\tilde {\Delta}_{\mathrm{ext}}}{\b\eta}(\quat^-, \b\omega^-, \b0) & \derp{\Delta_{\mathrm{ext}}}{\b\omega^-} \end{bmatrix}}_{\b D} \begin{bmatrix} \delta\b\xi^- \\ \delta\b\omega^- \end{bmatrix}.
\end{align}



\subsection{Geometric Model Predictive Control}

\begin{figure}[t!]
    \centering
    \includegraphics[width=0.95\columnwidth]{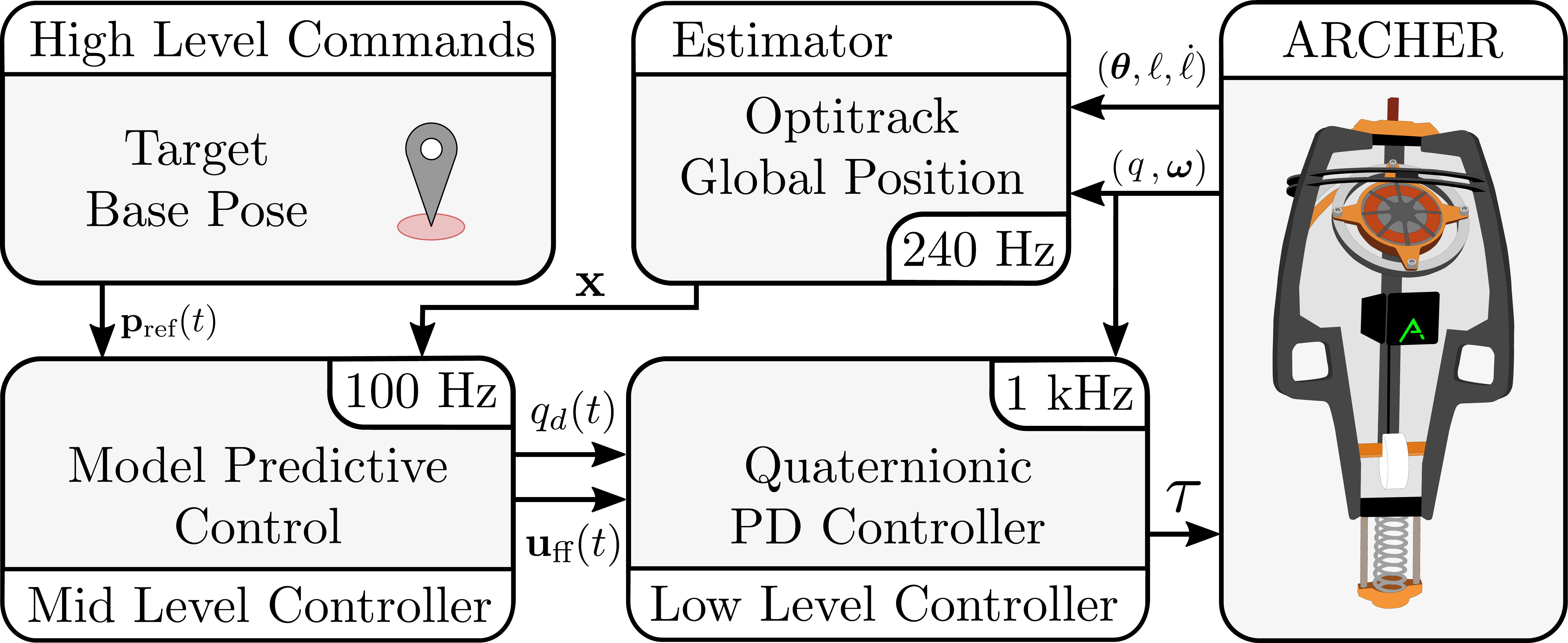}
    \caption{The multi-rate architecture employed for the robot.}
    \label{fig:MultiRate}
\end{figure}

This section represents the mid-level of the control hierarchy, as shown in \figref{fig:MultiRate}. High level target base positions $\b p_{\textrm{ref}} \in \R^3$ are provided by the user, and MPC produces reference trajectories to pass to the low level. This architecture maintains the benefit of having a horizon and is paired with a low level feedback controller which adds robustness to model error and delays induced by computation time. 

For a given MPC horizon $N\in \mathbb{N}$, we begin by constructing a vertex sequence $v_k \in V$ for $k=0,\ldots,N-1$ describing the continuous modes that the robot will be in at various points along the horizon. These are defined \textit{a priori} by estimating the time to impact of the robot. We also construct a sequence of $e_k = v_k\to v_{k+1}\in E\cup\{0\}$, where $e_k = 0$ if no discrete transition is expected. 
Consider a discrete (manifold valued) state trajectory $\bar{\b x}_k$ and input trajectory $\bar {\b{u}}_k$. We introduce the variables:
\begin{align*}
    \bar{\b z}_k = (\bar{\b p}_k, \bar{\b \xi}_k ,\bar{\b \theta}_k, \bar{\ell}_k, {\bar{\b v}}_k) \in \R^{20},
\end{align*}
with $\bar{\b\xi}_k$ defined as in \eqref{eqn:xi_def}, and whereby $\b z_k$ will represent our decision variables in the MPC program. At each index $k$, compute the linearizations of the dynamics in the vertex $v_k$:
\begin{align}
    \dot {\b z}_k = \b A_{v_k}\b z_k  &+ \b B_{v_k} \b u_k \notag\\
    &+ \underbrace{\b f_{v_k}(\bar {\b x}_k,\bar{\b u}_k) - \b A_{v_k}\bar {\b z}_k - \b B_{v_k} \bar {\b u}_k}_{\triangleq \b C_{v_k}}, \label{eqn:cont_lin_2}
\end{align}
where the $\quat_k$ and $\b\omega_k$ elements are linearized as in \eqref{eqn:cont_lin}, the Euclidean elements are linearized in the standard way.
From \eqref{eqn:cont_lin_2}, we can produce a discrete-time linear system over a time interval $h \in \R_{>0}$ by taking an Euler step (in Euclidean space), or by using the matrix exponential in $\mathbb{R}^{20\times 20}$ to produce the discrete time dynamics:
\begin{align}
    \z_{k+1} &= \b A^d_{v_k} \z_k + \b B^d_{v_k} \b u_k + \b C^d_{v_k},\label{eqn:cont_lin_disc}\\
    \z_k^+ &= \b D_{e_k} \z_k^- + \underbrace{\b \Delta_{e_k}(\bar {\x}_k^-) - \b D_{e_k}\bar {\z}_k }_{\triangleq \b E_{e_k}}.\label{eqn:disc_lin}
\end{align}
%
 We now introduce the finite-time optimal control problem (FTOCP), i.e., the \emph{geometric model predictive controller}:
\begin{subequations}
\begin{alignat}{2}
    \hspace{-1mm}\min_{\substack{\b{u}_{k}, \z_{k}}} \quad & \sum_{k=0}^{N-1} (\z_k - \z_{\textrm{ref}})^\top \b Q (\z_k - \b\z_{\textrm{ref}}) + \b u_k^\top && \b R\b u_k  + \z_N^\top \b V \z_N \notag\\
\textrm{s.t.}~ \quad & \z_{k+1} = \mb A^d_{v_k}\z_{k} + \mb B^d_{v_k}\b u_{k} + \mb C^d_{v_k}, && \textrm{if}~e_k = 0 \label{eqn:mpclindyn} \\
  & \z_{k+1} = \mb D_{e_k}\z_{k} + \mb E_{e_k},&& \textrm{if}~e_k \ne 0\label{eqn:mpclindyn} \\
  & \z_{0} = \z(t),&& \label{eqn:mpcic} \\
  & \b u_k\in \mathcal{U}&&
  \end{alignat}
\end{subequations}
where $\b Q \in \mathbb{S}_{>0}^{2n}$ and $\b R \in \mathbb{S}_{>0}^{2n}$ are symmetric, positive definite state and input gain matrices, respectively, $\b V \in \mathbb{S}_{>0}^{2n}$ is a quadratic approximation of the cost-to-go, $\mathcal{U}$ is an input constraint set, and where the initial condition $\b\xi_0 = \mb 0$ is enforced, as previously discussed. The above optimal control problem is solved in an SQP fashion, where the solution from the previous iteration is used to produce the linearizations for the next. Specifically, we can take $(\bar \z_k, \bar{\b u}_k) = (\b z_k^*, \b u_k^*)$ where the asterisk indicates the optimal solution, and $\bar \x_k$ can be produced from $\bar \z_k$ via inverting \eqref{eqn:xi_def}.

\subsection{Quaternionic Feedback}
Once MPC produces a solution, a desired trajectory and feedforward input can be produced as:
\begingroup
\setlength{\abovedisplayskip}{7pt}
\setlength{\belowdisplayskip}{9pt}
\begin{align*}
    \quat_d(\tau) &= \bar{\quat}_0\exp(\b \xi^*_1), &  \b \omega_d(\tau) &= \b\omega^*_1, &   \b u_{\textrm{ff}}(\tau) &= \b u^*_0,
\end{align*}
\endgroup
for $\tau \in [0,dt)\subset\R$. An interesting area of future work is using the MPC signals to produce dynamically admissible trajectories in the inter-MPC times, but this was not explored due to communication bandwidth limitations. The FTOCP is implemented in a receding horizon fashion, where the low level controller only ever receives the first control input and desired trajectory. 

\begin{figure}[!t]
    \centering
    \includegraphics[width=\columnwidth]{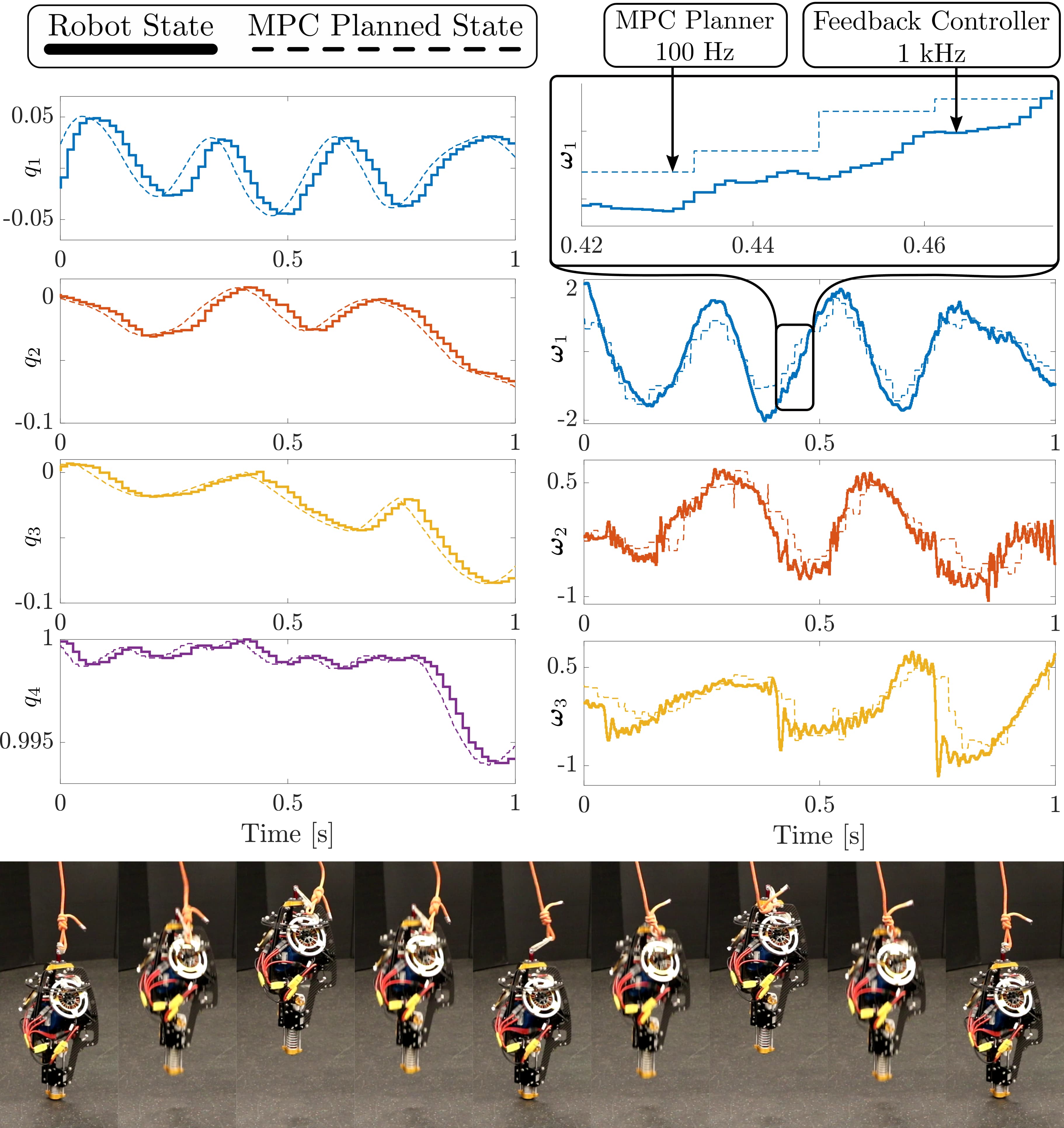}
    \caption{The planned elements $\quat\in S^3$ and $\b \omega\in\frak{s}^3$, as well as the low level feedback controller. The multi-rate nature of the methodology can be seen in the difference of time scales between when MPC produces trajectories and when the low level controller updates.}
    \vspace{1mm}
    \label{fig:hardData}
\end{figure}

Given the measured quaternion $\quat_a$, measured angular rate  $\b \omega_a$, desired quaternion $\quat_{d}$, and desired angular rate $\b \omega_d$ of the robot, we can construct our actuation as:
\begingroup
\setlength{\abovedisplayskip}{7pt}
\setlength{\belowdisplayskip}{9pt}
\begin{align*}
    \b u(\x, t) = -\b K_p  \mathbb{I}\textrm{m}({\quat_{ d}(t)^{-1} \quat_{ a}}) - \b K_d(\b\omega_{ a} - \b\omega_{ d}(t)) + \b u_{\textrm{ff}(t)},
\end{align*}
\endgroup
where $\b K_p, \b K_d\in \mathbb{S}_{>0}^3$ are positive definite gain matrices. The product $\quat_d(t)^{-1}\quat_a$ represents a ``difference" between elements of $S^3$; if $\quat_a$ is in a small neighborhood of $\quat_d$, then the product is in a small neighborhood of the identity element $\quat_{\mathcal{E}}$. The map $\mathbb{I}\textrm{m}:S^3\to \mathfrak{s}^3$ takes the purely imaginary component of the error signal, and can be viewed as the Euclidean projection of the Lie group onto the Lie algebra, allowing us to base the control input over a vector valued error. 
Alternatively, the $\log$ operation could be used instead of $\mathbb{I}\textrm{m}$, but the $\mathbb{I}\textrm{m}$ operator was empirically found to work more reliably. 

\section{Implementation and Results}
\label{sec:results}
\begin{figure}[t!]
    \centering
    \includegraphics[width=\columnwidth] {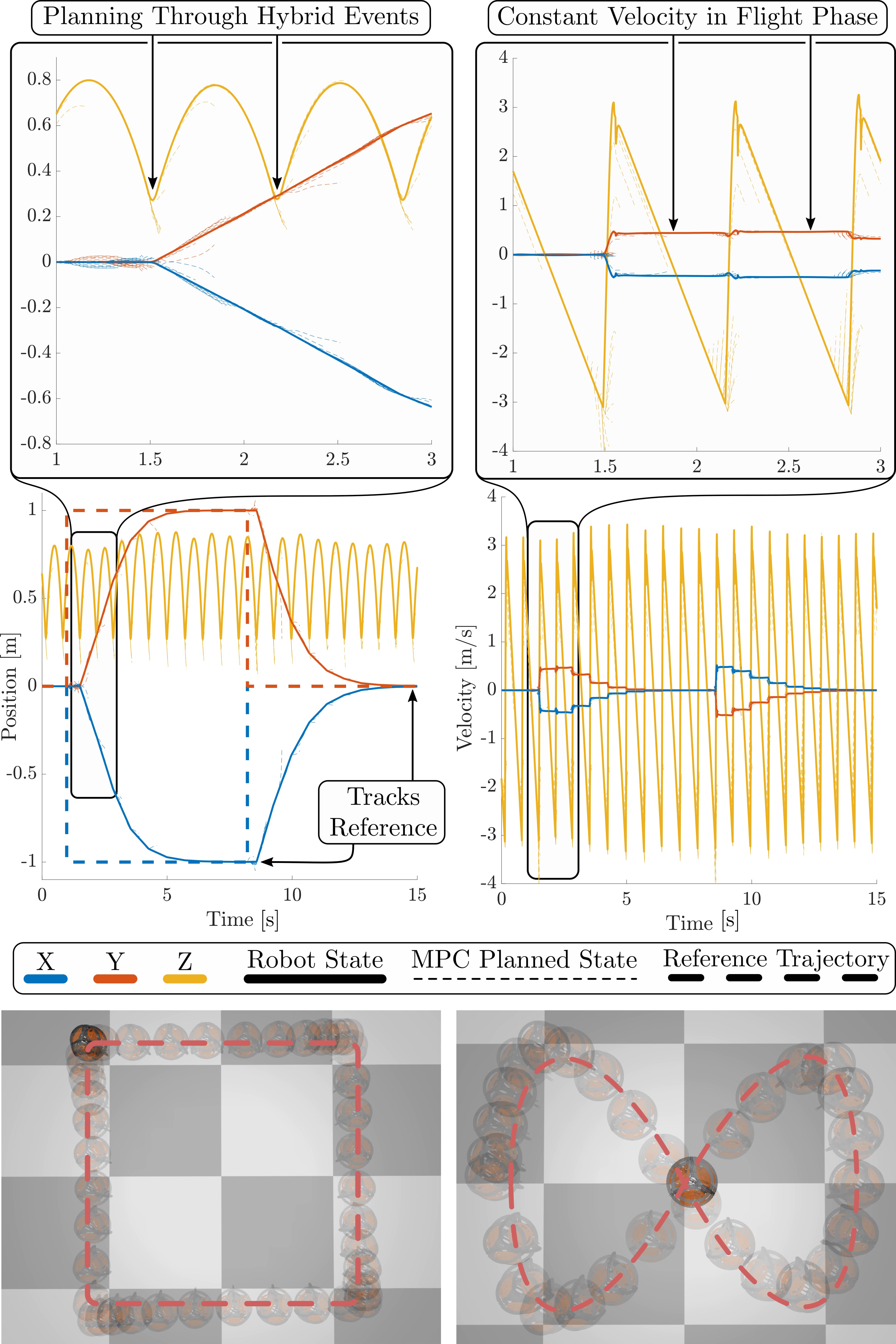}
    \caption{(Above) Positions and velocities of the robot tracking a global reference setpoint in simulation. Note the planning through hybrid events and constant velocity in flight phase due to underactuation. (Below) Two reference trajectories.}
    \label{fig:simTracking}
\end{figure}
\begin{figure*}
    \centering
    \includegraphics[width=0.95\textwidth]{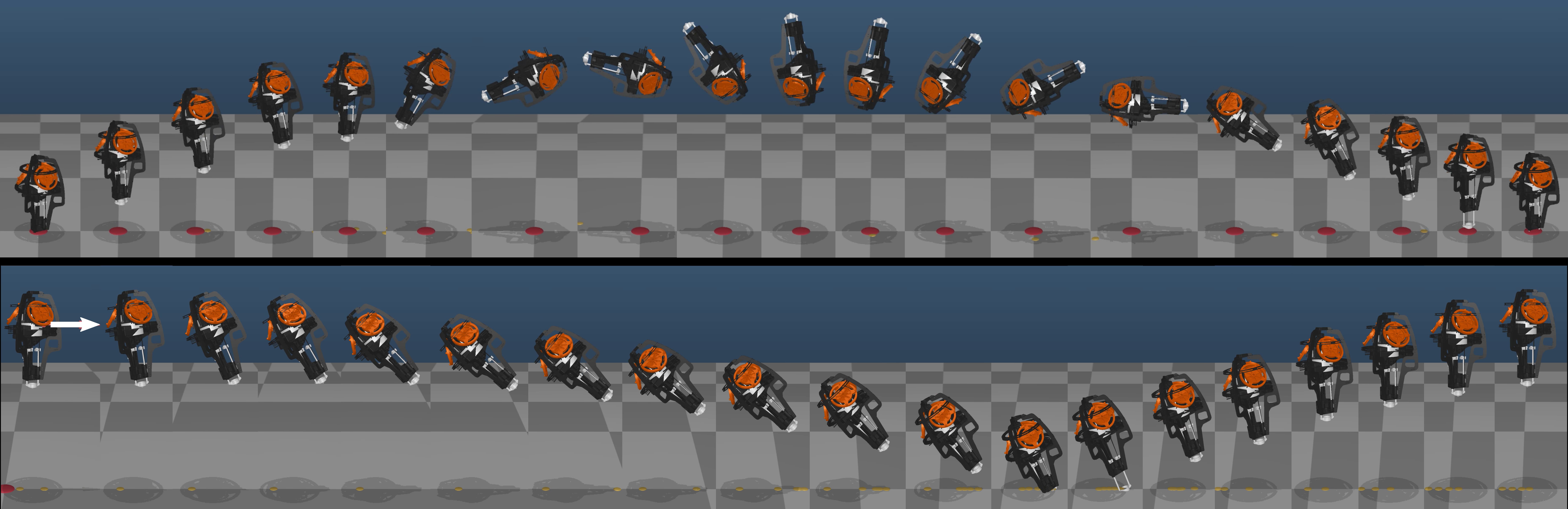}
    \caption{Dynamic motions explored in simulation, including flipping (above) and disturbance rejection (below).\vspace{-5mm}}
    \label{fig:simFun}
\end{figure*}
\subsection{Hardware}
The ARCHER \cite{ambrose_creating_2022} hardware platform consists of three KV115 T-Motors with 250 g flywheel masses attached for orientation control, and one U10-plus T-Motor attached to a 3-1 gear reduction to the foot via a cable and pulley system. The robot is powered by two 6 cell LiPo betteries connected in series, which can supply up to 50.8 V at over 100 A of current to the four ELMO Gold Solo Twitter motor controllers. The robot has two on-board Arduino Teensy microcontrollers for the low level feedback control, which run at 1kHz and communicate over WiFi via an ESP module to a desktop running the mid-level controller.

The MPC program runs at 100 Hz on an Ubuntu 20.04 machine with AMD Ryzen 5950x @ 3.4 GHz and 64 Gb RAM. The Pinocchio C++ library \cite{carpentier_pinocchio_2019} is used, specifically the pinocchio3 preview branch, to produce fast evaluation of the system dynamics (constrained, unconstrained, continuous, and discrete), as well as their associated Jacobians, and the manif C++ library \cite{sola_micro_2021} is used to handle all Lie group operations (such as $\log$ and $\exp$). As seen in \figref{fig:hardData} as well as the supplemental video \cite{noauthor_supplemental_video_2022}, the robot was successfully able to hop stably in place, demonstrating the first instance of 3D hopping using online motion planning.
\vspace{-1mm}
\subsection{Simulation}
In order to thoroughly test the method, the torque limits of the robot were increased in a Mujoco \cite{todorov_mujoco_2012} simulation environment from 1.5 Nm to 15 Nm. First, the tracking of various global reference signals, including a square and a Lissajous trajectory, were evaluated as seen in Figure \ref{fig:simTracking}. Note the constant global velocity in the flight phase due to the lack of control inputs when the robot is in the air. As such, the robot must carefully plan its interactions with the ground in order to track the desired reference signals. Specifically, it is interesting to see how MPC is implicitly able to control the actuated coordinates of the robot in order to stabilize the underactuated ones. Also note the discontinuities in the MPC planned trajectories around the impact events due to the hybrid nature of the system dynamics.
\begin{table}[b]
\begin{center}
\vspace{1mm}
\caption{MPC parameters} 
\begin{tabular}{| l | c || l | c| } 
 \hline 
 Horizon Length & 20 & SQP Iterations & 2 \\ \hline
 $\b p$ Weight & 10 &  $\b v$ Weight & 1 \\ \hline
 $\quat$ Weight & 10 &  $\b \omega$ Weight & 0.01 \\ \hline
 $\b u$ Weight & 0.001 & $\b u_{\textrm{max}}$  & 1.5 Nm\\ \hline
 $K_p$ Roll/Pitch Gain & 120 &$K_d$ Roll/Pitch Gain & 4 \\ \hline
 $K_p$ Yaw Gain & 15 &$K_d$ Yaw Gain & 1 \\ \hline
 $dt_{\textrm{flight}}$ & 0.01 & $dt_{\textrm{ground}}$ & 0.001 \\ \hline
\end{tabular}
\label{tab:param}
\vspace{-2mm}
\end{center}
\end{table}
Next, disturbance rejection and more dynamic maneuvers like flipping were tested on the system, as seen in the accompanying video \cite{noauthor_supplemental_video_2022} and \figref{fig:simFun}. Due to the geometrically consistent structure of the planning algorithm, the robot is able to explore a variety of states on its orientation manifold, and exhibits exceptional robustness to disturbances. Note that the torque limitations, but not limitations of the methodology, prohibit such demonstrations on the hardware platform.
\vspace{-1mm}
\subsection{Implementation Details}
The complete list of parameters used in the MPC program are detailed in \tabref{tab:param}. There is an inherent tradeoff between tracking global position and maintaining a vertical orientation -- as such the associated gains need to be appropriately tuned. To avoid adding a nonlinear and mixed integer constraint in to the optimization program, the impact time was calculated as though the hopper was exactly vertical via solving for the ballistic trajectory in the $z$-direction. As the spring dynamics add significant stiffness to the optimization problem, the foot torque was set to zero in the optimizer, and instead the MPC program plans as though it is a passive degree of freedom. Instead, the low level controller runs its own feedback controller to regulate foot compression between impact events. As the dynamics of the ground phase are more challenging than the flight phase, the system was discretized more finely in that domain. This means that the lookahead time shrank whenever impact came into view in the horizon, as the horizon length was kept constant.  We found that using the matrix exponential instead of the Euclidean Euler step aided in performance, likely due to the reduced one step prediction error. Finally, the quadratic approximation of the cost to go was simply taken to be $\b Q$. The complete code can be found at \cite{noauthor_complete_2022}. 



\section{Conclusion and Future Work}
In this work, we demonstrated that predictive control is successfully able to regulate the underactuated coordinates of a 3D hopping robot through intelligent control of the actuated states. Through the use of geometrically consistent model predictive control and feedback layers, we were able to achieve stable hopping on hardware, and trajectory tracking and flipping in simulation. Future work includes providing theoretical justification of why predictive control is able to regulate the underactuated coordinates of robotic systems, as well as a proof of state and input constraint satisfaction in continuous time for underactuated systems. On the hardware side, planning with constrained footholds, incorporating friction cone constraints, and the inclusion of a high level decision making layer will be investigated.

\section{Acknowledgements}
The authors would like to especially thank Eric Ambrose for providing us with such a well-built hardware platform, and for supporting our low-level controller implementation efforts. We would also like to sincerely thank Igor Sadalski, as well as Sergio Esteban and Adrian Boedtker Ghansah for their help with simulation and hardware implementation, and Will Compton for his experimental assistance. 

\newpage\newpage 
\bibliographystyle{IEEEtran}
\balance
\bibliography{main}

\end{document}